\documentclass[conference]{IEEEtran}
\usepackage{cite}
\usepackage{amsmath,amssymb,amsfonts}
\usepackage{algorithmic}
\usepackage{graphicx}
\usepackage{textcomp}
\usepackage{xcolor}
\usepackage{subcaption}
\def\BibTeX{{\rm B\kern-.05em{\sc i\kern-.025em b}\kern-.08em
    T\kern-.1667em\lower.7ex\hbox{E}\kern-.125emX}}

\begin{document}

\title{Testing RadiX-Nets: Advances in Viable Sparse Topologies}

\author{\IEEEauthorblockN{Kevin Kwak, Zack West, Hayden Jananthan, Jeremy Kepner}
\IEEEauthorblockA{\textit{Massachusetts Institute of Technology}}}

\maketitle

\begin{abstract}
    The exponential growth of data has sparked computational demands on ML research and industry use. Sparsification of hyper-parametrized deep neural networks (DNNs) creates simpler representations of complex data. Past research has shown that some sparse networks achieve similar performance as dense ones, reducing runtime and storage. RadiX-Nets, a subgroup of sparse DNNs, maintain uniformity which counteracts their lack of neural connections. Generation, independent of a dense network, yields faster asymptotic training and removes the need for costly pruning. However, little work has been done on RadiX-Nets, making testing challenging. This paper presents a testing suite for RadiX-Nets in TensorFlow. We test RadiX-Net performance to streamline processing in scalable models, revealing relationships between network topology, initialization, and training behavior. We also encounter ``strange models'' that train inconsistently and to lower accuracy while models of similar sparsity train well.
\end{abstract}


\section{Introduction} \label{intro}

\let\thefootnote\relax\footnotetext{Research was sponsored by the United States Air Force Research Laboratory and the Department of the Air Force Artificial Intelligence Accelerator and was accomplished under Cooperative Agreement Number FA8750-19-2-1000. The views and conclusions contained in this document are those of the authors and should not be interpreted as representing the official policies, either expressed or implied, of the Department of the Air Force or the U.S. Government. The U.S. Government is authorized to reproduce and distribute reprints for Government purposes notwithstanding any copyright notation herein. $(979-8-3503-0965-2/23/$31.00 ©2023 IEEE)}

Deep neural networks have been employed in AI industries far removed from traditional computer science applications. Research has shown that increasing neural network size and depth is sure to lead to better performance \cite{pmlr-v49-eldan16}. The current trend is to use hyper-parametrized models, often exceeding billions of parameters. Thus, the requirement for matrix multiplication and other computational procedures essential for neural network training is consistently growing. 

One way to decrease memory requirements in neural networks is to introduce sparsity among connections \cite{Kepner_Jeremy2018-07-17}. 
This approach finds its roots in biology, inspired by the human brain's remarkable sparsity ($\approx 2 \times 10^{-8}$) in synaptic connections between neurons \cite{Azevedo2009-qu}. 
The field of model compression mostly focuses on connection pruning, where large models are trained and subsequently pruned to a fraction of the original size with almost no drop in accuracy. Although this method is effective for compressing models for deployment purposes, its training efficiency falls short due to the models' initial density. Not much research has been done on neural networks which adhere to a sparse topology throughout the entire training process. Additionally, the sparse neural network structures that have been researched seem to have performed worse than models found by pruning \cite{evci2020difficulty}.

Given the limited research on initially sparse neural network topologies, the concept of RadiX-Nets was introduced in \cite{kepner2019radix}. RadiX-Nets take their inspiration from mixed-radix number systems composed of overlapping decision tree unions (see Figure~1 of \cite{kepner2019radix} for an explicit illustrated example) and X-Nets \cite{prabhu2018deep}. Experimental research on these topologies exhibits the presence of inconsistent training concavity and performance across topological initializations \cite{alford2019training}.

\S\ref{setup and training section} describes the philosophy and details behind the creation of the RadiX-Net Testing Suite which facilitates the examination of training behaviors for initially sparse deep neural networks (with an emphasis on RadiX-Nets) by providing a user-friendly and comprehensible implementation of RadiX-Nets, featuring customizable TensorFlow classes and supporting functions for RadiX-Net generation and visualization tools. 
In \S\ref{results section} the RadiX-Net Testing Suite is leveraged to examine training behaviors exhibited by RadiX-Nets using the Lenet 300-100 architecture \cite{lecun1998gradient} and MNIST dataset \cite{deng2012mnist}, particularly assessing the impact of random weight initialization, sparsity, and varying radices (which determine the topological structure) on the creation and performance of these models. Among these results are the findings that the performance of RadiX-Nets is not guaranteed: instances of chaotic phenomena emerge as sparsity levels increase, albeit intermittently. Specifically, the rise in frequency (with respect to sparsity) of what we call ``strange models'',  which train sporadically. Random initializations also affect the ability of a model to learn. The existence of ``strange models'' suggests that, in generating RadiX-Nets, radices might hold more significance than sparsity under certain circumstances.

\section{Setup and Training} \label{setup}
\label{setup and training section}

\subsection{MIT SuperCloud setup}

The RadiX-Net Testing Suite was made and run on the MIT SuperCloud, which partitioned two Intel Xeon-P8 CPU nodes and 8 gigabytes of memory. It also ran with Python version 3.9.16 \cite{van2009python} and TensorFlow version 2.11.1 \cite{mart2015tensorflow}.

All of the plots in \S \ref{results-section} were generated with the MIT SuperCloud, with just one CPU partitioned from Xeon-P8 and with 4 gigabytes of memory. 

Plotting was performed with NetworkX 2.8.8 \cite{hagberg2008exploring} and Matplotlib 3.7.1 \cite{hunter2007matplotlib}.

\subsection{RadiX-Net Testing Suite}

In order to facilitate research on RadiX-Nets, an easy to use and understandable implementation of RadiX-Nets in the RadiX-Net Testing Suite was created. Among the functions and classes making up the Testing Suite is the \verb`Mask` function, which allows the user to manually set a desired mask---which is applied at every iteration during model training to enforce the desired sparse topology---through a NumPy array called \verb`layerval`. The backbone of the RadiX-Net implementaiton is the \verb`RadixLayer` class; the \verb`RadixLayer` class takes in the same \verb`layerval` variable and trims it to fit the desired network structure. The \verb`RadixLayer` class also supports arbitrary activation functions to be passed to the underlying TensorFlow libraries. A \verb`CustomModel` class allows users to customize the number of layers and which radices are used to determine the RadiX-Net sparse topology, after which helper functions modify the shapes of the layer masks to fit the desired model shape and size.

\begin{figure}[htbp]
    \centering
    \includegraphics[scale=0.5]{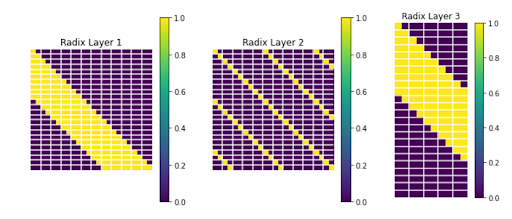}
    \caption{Heatmap adjacency matrix visualization tool in the RadiX-Net Testing Suite.}
    \label{fig:adj}
\end{figure}

\begin{figure}[htbp]
    \centering
    \includegraphics[scale=0.4]{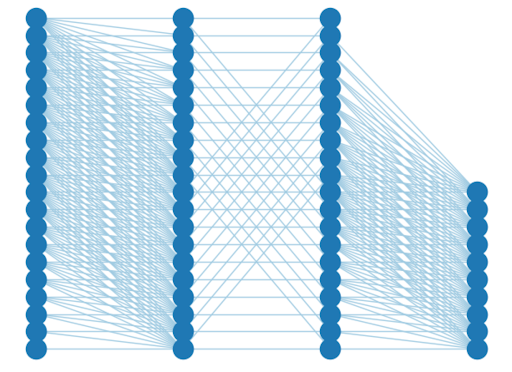}
    \caption{Graphical visualization tool in the RadiX-Net Testing Suite.}
    \label{fig:graphical}
\end{figure}

Along with extensive documentation, the RadiX-Net Testing Suite also provides visualization tools to help facilitate a better understanding of the structure of RadiX-Nets and the resulting sparse neural networks. The first of these tools is a heatmap adjacency matrix visualization, an example of which is shown in Figure~\ref{fig:adj}. Yellow-colored squares indicate nonzero weights (i.e., connections between neurons), while purple-colored squares signify zero weights (i.e., lack of connections between neurons). A more direct visualization of the sparse network topology is depicted in Figure~\ref{fig:graphical}, directly showing what connections exist between each layer of the network. A benefit of this approach is that it reveals the symmetrical nature of RadiX-Nets. Both visualizations are of a RadiX-Net created with radix list \verb`[[10, 10], [10]]` applied to the Lenet-300-100 architecture.

While the previous components of the RadiX-Net Testing Suite which constructed mainly as educational tools, the Testing Suite includes additional tools for RadiX-Net testing and applications, including functions comparing training behaviors between initially sparse networks and post-training pruned netowrks (like the analyzes originally documented in \cite{alford2019training}) and functions comparing training behaviors and accuracies of RadiX-Nets across sparsity levels, seeds, and radix lists. Plots illustrating these functions are seen in \S\ref{results section}. Additionally, pytest unit tests are included to ensure validity and consistency of the Suite.

As previously mentioned, all plots and results shown in \S\ref{results section} were run with the Lenet 300-100 architecture on the MNIST dataset.

\section{Results} \label{results-section}
\label{results section}

\begin{figure}[htbp] separate page)
    \centering
    \includegraphics[scale=.5]{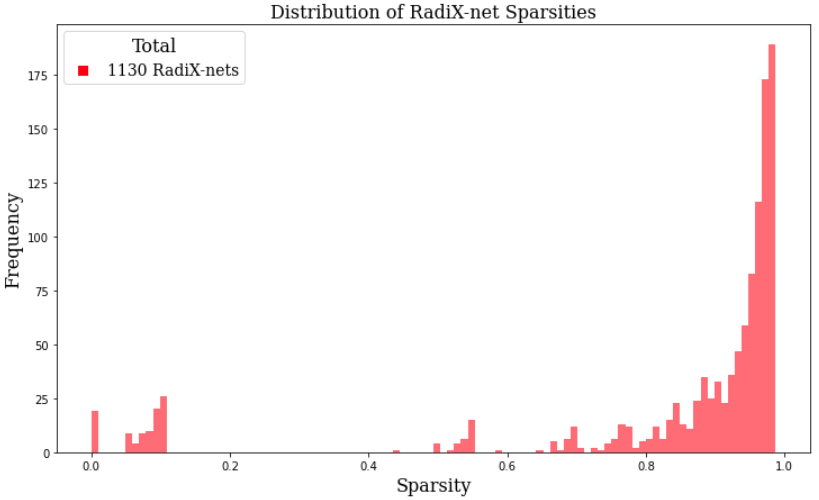}
    \caption{Distribution of RadiX-Net sparsities for valid permutations of factors 1--10.}
    \label{fig:distribution}
\end{figure}

Figure~\ref{fig:distribution} shows the skew of RadiX-Net generation towards higher sparsities. This is due to the multiplicative nature of RadiX-Net sublists \cite{kepner2019radix}---a positive correlation exists between the product of a radix sublist and the sparsity of the RadiX-Net. This distribution skew is ideal for targeting asymptotically large networks because most topologies result in high sparsity.

\begin{figure}[htbp]
    \centering
    \includegraphics[scale=.5]{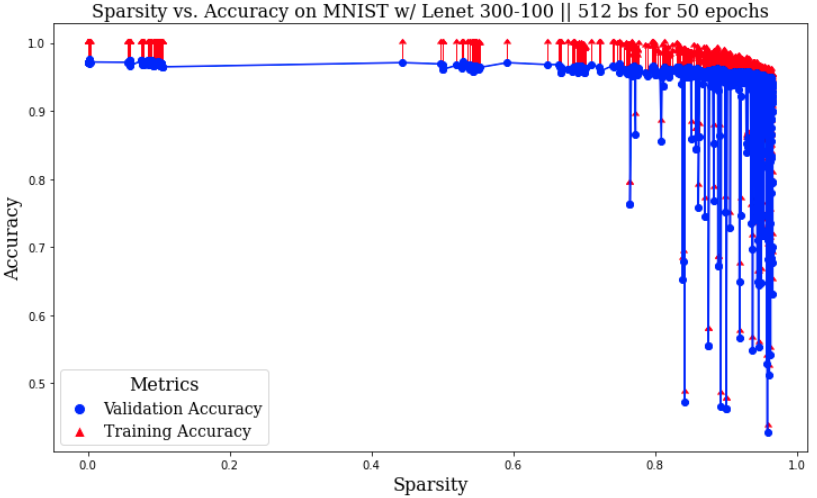}
    \caption{Accuracy metrics for RadiX-Nets generated by valid permutations of factors from 1 to 10.}
    \label{fig:acc-met} 
\end{figure}

Figure~\ref{fig:acc-met} displays the training (red chevrons) and validation accuracies (blue dots) of 1130 distinct RadiX-Nets. These networks were initialized from legal radix lists (see \cite{kepner2019radix} for radix list requirements) that permute factors and sublists of integers from 1 to 10. Once the sparsity level surpasses 75\%, there is an increasing occurrence of erratic training behavior, particularly at higher sparsity levels, illustrating the high variance among RadiX-Net topologies even when accounting for sparsity.
The inherent structure of a RadiX-Net ensures path-connectedness, leading to equidistant paths between input and output nodes. Therefore, this unstable training can be attributed to connection subsets formed by radices where a concentration of important input nodes have paths to a nonuniform set of output nodes, an effect which is accentuated in sparse networks.

\subsection{Random Initialization}

\begin{figure}[htbp]
    \centering
    \includegraphics[scale=.4]{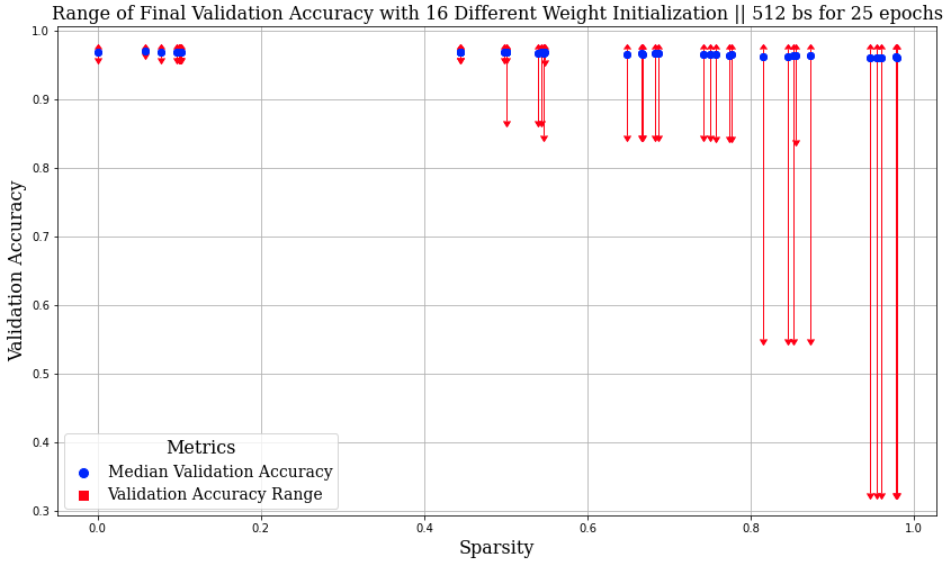}
    \caption{Range of validation accuracies when training different sparsity RadiX-Nets with varying seeds.}
    \label{fig:rangeseed}
\end{figure}

Figure~\ref{fig:rangeseed} represents the effect of different random weight initializations on the post-training accuracy of RadiX-Nets. For each sparsity classification (0--10\%, 10--20\%, \ldots, 90--100\%), five random networks were chosen from the distribution in Figure~\ref{fig:distribution}. Each topology was initialized and trained on sixteen different seeds; error bars represent the range of validation accuracies and the blue dot the median of the sixteen training runs. For denser networks, ranges are within 2\%, though a quick fluctuation occurs once sparsities exceed 50\%, increasing markedly in networks of 90\% sparsity or higher. However, across all sparsities, the median (blue dot) training accuracy remains stable, indicating that fewer connections can lead to a cascading effect on RadiX-Net consistency.

\begin{figure}[htbp]
    \centering
    \includegraphics[scale=.5]{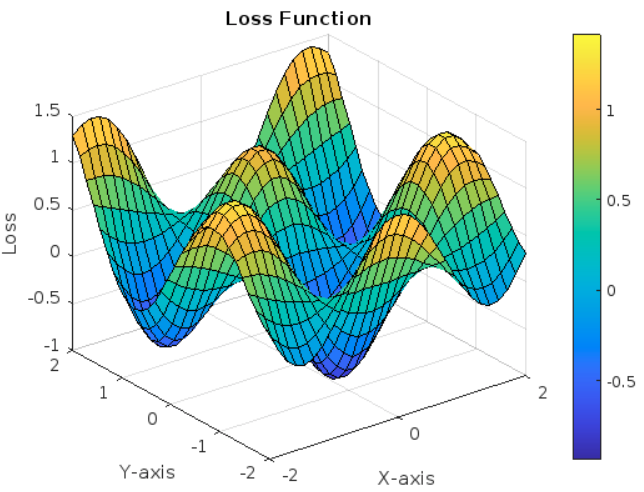}
    \caption{Representation of a 3-dimensional loss landscape via MATLAB \cite{higham2016matlab}}
    \label{fig:lossFunc}
\end{figure}

Initializing a neural network's weights essentially defines its position in a high-dimensional loss landscape. Different initializations represent different starting points and potential optimization trajectories.
Higher sparsity means fewer initial parameters, resulting in a lower-dimensional graph, or fewer ways for an optimizing function to traverse towards higher-optimality local minima. Reducing the number of connections leads to more deterministic initializations and could potentially limit optimizers' capability to explore a broader range of local minima.
In reference to Figure~\ref{fig:lossFunc}, this process can be visualized as a stochastic gradient descent towards a minimum in a 3-dimensional Cartesian coordinate system.
As dimensionality decreases from a highly-parametrized dense network, there exist fewer combinations of input weight combinations that minimize the loss function and thus maximize accuracy. This is crucial as it forms the basis of the Lottery Ticket Hypothesis \cite{frankle2018lottery}, and how RadiX-Nets and other sparse topologies can systematically drive convergence towards topologies by reducing dimensionality.

\subsection{Visualizing Uniformity}

\begin{figure}[htbp]
    \centering
    \includegraphics[scale=.4]{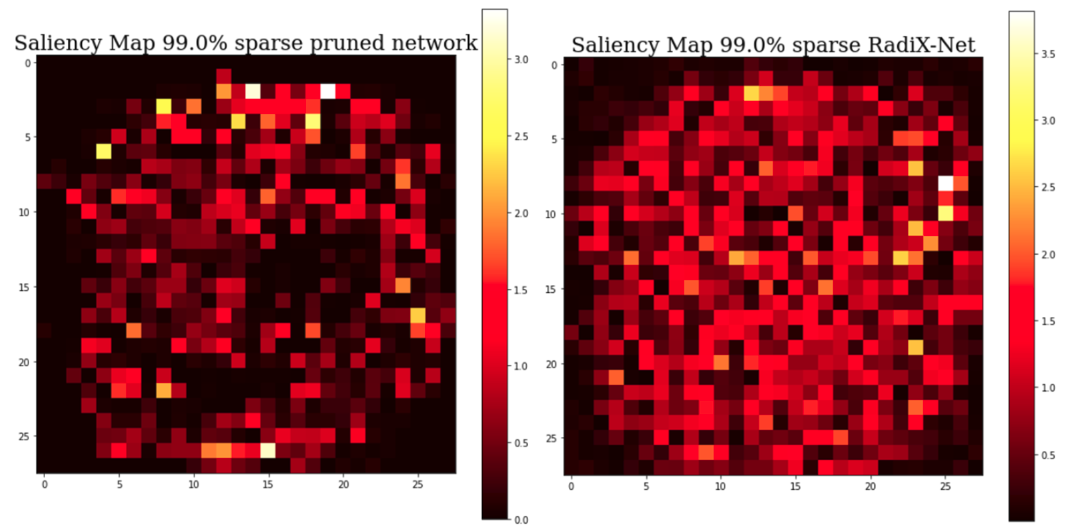}
    \caption{Saliency maps for a sparsity level of 99$\%$ in both low-magnitude pruned and RadiX-Net models trained on MNIST, respectively.}
    \label{fig:saliency}
\end{figure}

A saliency map \cite{mundhenk2020efficient} highlights important regions or features in input data, particularly those that are visually represented, such as images. Figure~\ref{fig:saliency} displays saliency maps for 99$\%$ sparsity in both low-magnitude-pruned and RadiX-Net models, obtained by aggregating outgoing connection weights of input pixels in each network. Low-magnitude pruning crudely removes weights that contribute little to the network, evident by the lack of activity in edge inputs. This absence occurs because inputs in the MNIST dataset \cite{deng2012mnist} tend to centralize information in inner pixels, causing similar background regions to lack unique information for distinguishing inputs. This leads to gradients minimizing the weight of those outer connections, and hence, compressing the model to these inner connections.

Conversely, the radial symmetry of the RadiX-Net's saliency map underscores its versatility across datasets and capacity to preserve information at high sparsities. This supports the idea that these networks' connectivity and symmetry properties have advantages over other sparsification techniques such as low-magnitude pruning, especially as network depth and sparsity increases \cite{BadrinarayananM15}. For increasingly deep networks, the equality of outgoing connections from nodes created by mixed radix topologies prevents information loss that is common in imbalances via low-magnitude pruning. These imbalances are fairly exclusive to magnitude-based pruning, as connections are removed without reference to potential future branching throughout deeper layers \cite{liu2021exploring}. Other network compression algorithms like Fisher pruning take into account the sensitivity of weight removal with second derivative matrices but are costly \cite{theis2018faster}. RadiX-Nets' minimal computation requirement and information retention due to uniformity create a strong medium between these options.

\subsection{Strange models}

Because high sparsity networks are of highest interest, we focused observation of RadiX-Nets with sparsity at least 90\%. 426 such models were built on the Lenet 300-100 architecture and the MNIST dataset using radices ranging from 1 to 15. While most models trained relatively well---seeing accuracy near 90\% accuracy after only three epochs with batch size 100---several models exhibited vastly different training behavior, effectively the opposite of lottery ticket models \cite{frankle2018lottery}, which we term ``strange models'' for the remainder of the paper. 

\begin{figure}[htbp]
    \centering
    \vspace{-1cm}
\includegraphics[scale=0.38]{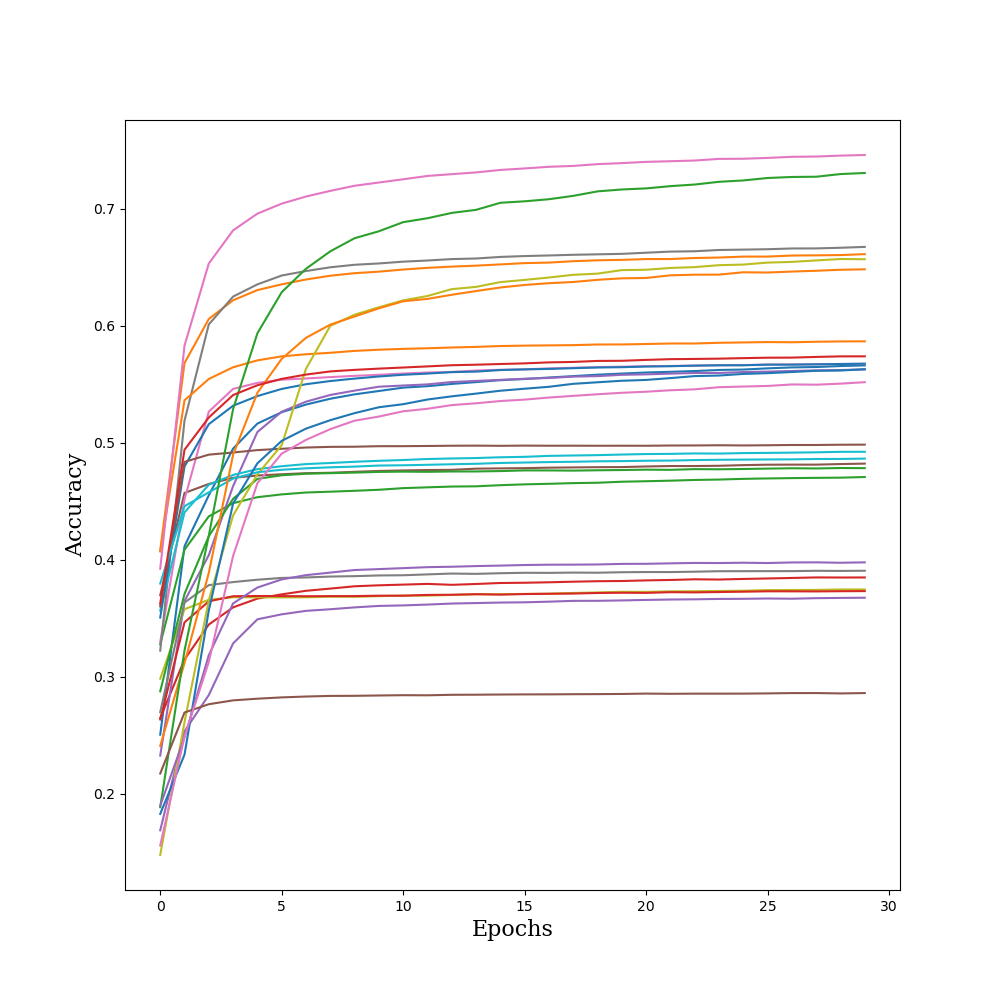}
    \caption{All found strange models' training behavior over 30 epochs.}
    \label{fig:allstrange}
\end{figure}

Figure~\ref{fig:allstrange} shows the training behavior of all strange models found out of the original 426 models, showing highly variable accuracies that fail to match the `normal' models after an order of magnitude more epochs (30 for the strange models versus 3 for the `normal' models). This shows that trained models with close sparsities need not have close model accuracies, and suggests that the specific radices chosen can be more deterministic of model accuracy. An additional observation supporting this last claim is that each of the strange models had at least one radix of 10 or higher. 

\begin{figure}[htbp]
    \centering
    \includegraphics[scale=0.43]{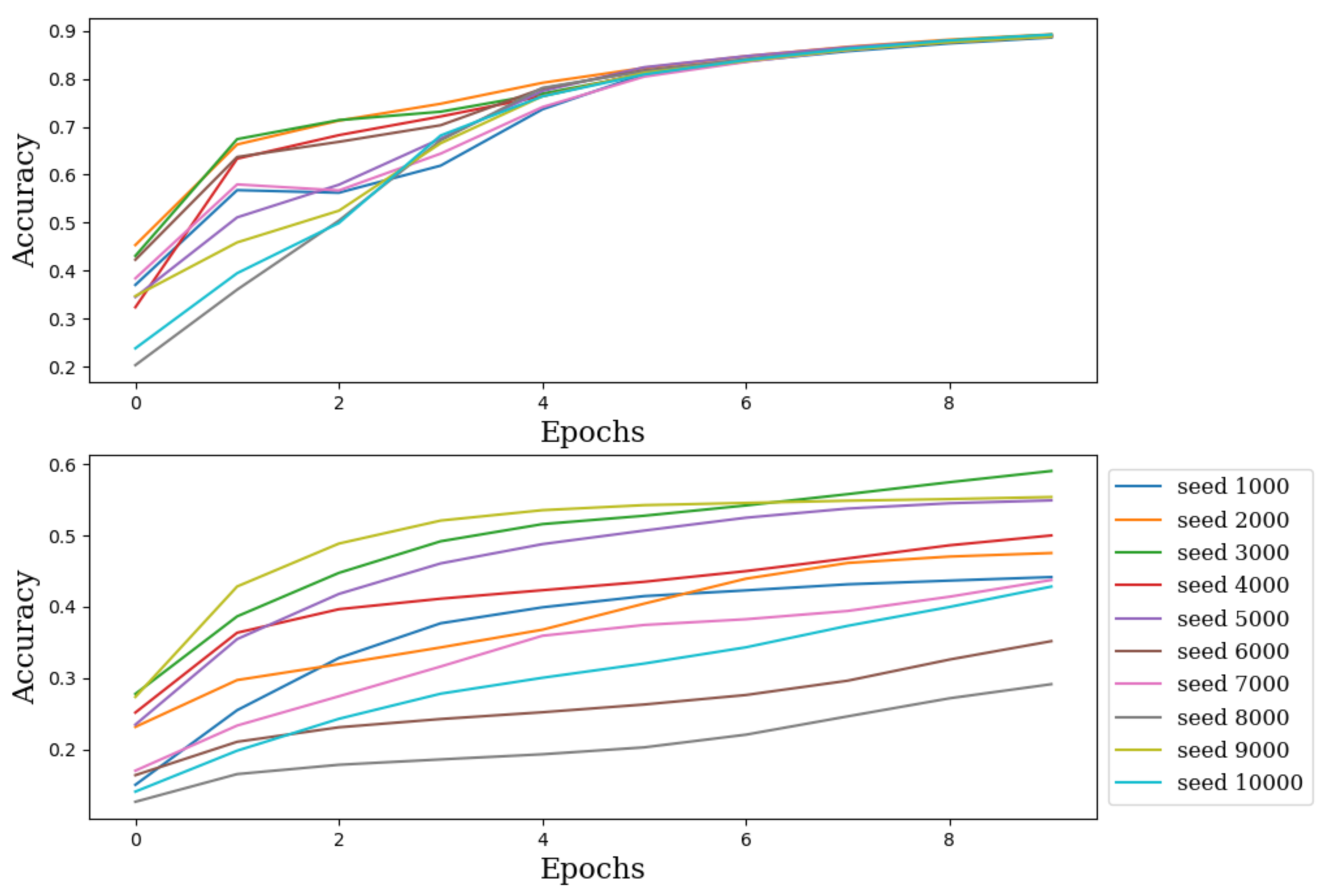}
    \caption{The effect of changing the seed on a normal model (top) and strange model (bottom).}
    \label{fig:seedeffect}
\end{figure}

As seen in Figure~\ref{fig:allstrange}, strange models' accuracies plateau quickly, implying that they---at least with their random initializations---get stuck in highly nonoptimal local minima early in training. Another peculiarity is that these models all seem to converge to different accuracies ranges that are fairly consistently separated. In Figure~\ref{fig:allstrange} 6 groupings of these models are presents with accuracies around 28\%, 35\%--40\%, 45\%--50\%, 55\%--57.5\%, 60\%--65\%, and 70\%--80\%. Whether the model accuracies of the strange models falling into well-separated ranges generalizes to other model architectures, datasets, and ranges of radices remains an open question. 
To further examine the effect of initialization on strange models, Figure~\ref{fig:seedeffect} shows the training behavior of a `normal' model (top) versus a strange model (bottom) under different initializations (seeds of 1000, 2000, \ldots, 10000). Whereas the different initializations of the `normal' model ultimately converged in greater than 90\% accuracy, the same could not be said about those of the strange model which both stayed separated and below 60\% over ten epochs. This suggests the behavior seen in Figure~\ref{fig:rangeseed} is largely attributable to these so-called strange models and that the effect of initialization on training accuracy of RadiX-Nets depends highly on the particular radices chosen.
In relation to the Lottery Ticket Hypothesis \cite{frankle2018lottery}, which roughly states that dense, randomly initialized, feed-forward networks contain performant subnetworks (when trained in isolation), our observations suggest that `winning tickets' can be found among the `normal' models.

\section{Conclusion}

This paper presents the RadiX-Net Testing Suite which serves to expedite future research in sparse topologies for deep neural networks. The Testing Suite is easily customized and contains educational visualization tools, thorough documentation, and feature model comparisons. Tools are included for comparing training behaviors of initially sparse networks and pruned networks as well the effects of sparsity, initialization, and varying radix list permutations on model learning.

The experimental results show a strong introduction for RadiX-Nets as a viable sparse topology. It is evident that certain radices generate sparse topologies that match the validation metrics of their dense counterparts. The erratic learning of RadiX-Nets generated by other radices, as originally documented in \cite{alford2019training}, becomes pronounced primarily at hyper-sparse levels, but remains an infrequent phenomenon before those sparsities and is not indicative of most RadiX-Nets' performance.

Future work should focus on creating heuristics for which radices correlate to this irregular learning. Fundamentally understanding the underlying causes of this instability and implementing preventive measures will consistently yield high performance with minimal computation, ideally addressing the problem of big data with respect to machine learning scalability. Doing so may require a more tuned selection of integers to permute in radix lists. Furthermore, testing the reliability of promising RadiX-Nets across different data and network architectures is necessary for the dependable employment of machine learning models.

\section*{Acknowledgment}

The authors wish to acknowledge the following individuals for their contributions and support: S. Alford, W. Arcand, W. Bergeron, D. Bestor, C. Birardi, B. Bond, S. Buckley, C. Byun, G. Floyd, V. Gadepally, D. Gupta, M. Houle, M. Hubbell, M. Jones, A. Klien, C. Leiserson, K. Malvey, P. Michaleas, C. Milner, S. Mohindra, L. Milechin, J. Mullen, R. Patel, S. Pentland, C. Prothmann, A. Prout, A. Reuther, R. Robinett, A. Rosa , J. Rountree, D. Rus, M. Sherman, C. Yee.

\bibliographystyle{IEEEtran}
\bibliography{references}

\end{document}